\documentclass[letterpaper, 10 pt, conference]{ieeeconf}  

\IEEEoverridecommandlockouts                              
\usepackage{graphics} 
\usepackage{epsfig} 
\usepackage{mathptmx} 
\usepackage{times} 
\usepackage{amsmath} 
\usepackage{amssymb}  

\usepackage{mathtools}          
\usepackage{mathrsfs}           
\usepackage{graphicx}           
\usepackage{subcaption}         
\usepackage[space]{grffile}     
\usepackage{url}                
\usepackage{lipsum}             

\usepackage{booktabs}
\usepackage{algorithm}
\usepackage{algpseudocode}

\title{\LARGE \bf
RANDPOL: Parameter-Efficient End-to-End Quadruped Locomotion via Randomized Policy Learning
}

\author{Zhuochen Liu, Rahul Jain, Quan Nguyen
\thanks{Zhuochen Liu and Rahul Jain are with the Ming Hsieh Department of Electrical and Computer Engineering, University of Southern California,
        Los Angeles, CA 90089, USA
        {\tt\small \{liuzhuoc,rahul.jain\}@usc.edu}}%
\thanks{Quan Nguyen is with the Department of Aerospace \& Mechanical Engineering, University of Southern California,
        Los Angeles, CA 90089, USA
        {\tt\small quann@usc.edu}}%
}

\begin{document}

\maketitle
\thispagestyle{empty}
\pagestyle{empty}

\begin{abstract}
Modern learning-based locomotion controllers typically rely on fully trainable deep neural networks with a large number of parameters. This paper studies a different design point for end-to-end control: whether effective quadruped locomotion can be achieved with a drastically reduced trainable parameter space. We present RANDomized POlicy Learning (RANDPOL), a policy learning approach in which the hidden layers of the actor and critic are randomly initialized and fixed, while only the final linear readout is trained. This yields a parameter-efficient controller class that retains nonlinear expressiveness through a fixed random basis while substantially reducing the dimension of the optimization problem. RANDPOL is supported by the mathematical foundation of randomized function approximation, which provides a principled basis for using fixed random nonlinear features as expressive function classes. We evaluate RANDPOL on end-to-end locomotion control for the Unitree Go2 quadruped and compare it with Proximal Policy Optimization (PPO). The results show that RANDPOL attains comparative locomotion performance with far fewer trainable parameters, lower learning-phase computation time per iteration, and a favorable performance-complexity trade-off. We further demonstrate successful zero-shot sim-to-real transfer of the learned RANDPOL controller on the physical Unitree Go2 under user-issued forward-velocity and yaw-rate commands. These results indicate that, for structured robotic control problems, reducing trainable complexity can remain compatible with effective simulated and real-world performance.
\end{abstract}

\section{Introduction}
\label{sec:introduction}

Many control problems can be formulated as sequential decision-making tasks and modeled as Markov Decision Processes (MDPs). In continuous-state, continuous-action settings with unknown dynamics, reinforcement learning (RL) provides a practical framework for learning feedback policies directly from data. This has enabled strong results in challenging domains such as games and legged locomotion, especially when large-scale simulation can supply abundant training experience \cite{c1, c2, c3, c4, c5, c6, c7, c8, c9, c10, c11}.

A dominant trend in modern RL is to use increasingly expressive neural network policies and value functions with a large number of trainable parameters. While this often improves representation power, it also increases optimization complexity, hyperparameter sensitivity, and training variance. In control applications, this raises a basic question: can effective end-to-end control still be achieved when the trainable component is made much smaller?

This paper studies that question through \emph{RANDomized POlicy Learning} (RANDPOL), a policy learning approach based on randomized neural networks. In RANDPOL, hidden layers are randomly initialized and kept fixed, while only the final layer is trained. As a result, the number of trainable parameters is drastically reduced, and policy learning becomes the problem of fitting a linear combination of randomly generated nonlinear features. This construction is principled rather than ad hoc: randomized neural networks have a well-established mathematical foundation and can be interpreted as learning over a fixed random basis \cite{c12, c13, c14}. RANDPOL builds on this foundation and adapts it to continuous-control policy learning.

The central perspective of this paper is that, for structured control problems, \emph{randomization is cheaper than optimization}. Instead of repeatedly updating every parameter of a deep network end-to-end, one can generate a rich fixed random basis and optimize only a small trainable readout. This yields a controller class with far fewer trainable parameters, a simpler optimization structure, and a smaller hyperparameter design space, while still retaining effective nonlinear control behavior.

We evaluate this idea on quadruped locomotion with the Unitree Go2 robot. Specifically, we use RANDPOL to train an end-to-end locomotion controller in simulation and demonstrate zero-shot sim-to-real transfer on hardware. Our main empirical comparison is against Proximal Policy Optimization (PPO) \cite{c15}, a widely used on-policy baseline with fully trainable neural networks. Because PPO uses a much larger trainable model class, our focus is not on absolute dominance in return, but on the trade-off between performance and trainable complexity. We show that RANDPOL attains comparative locomotion performance with far fewer trainable parameters, a simpler model class, and lower learning-phase computation time per iteration. The experiments further show that RANDPOL exhibits lower variability in mean reward across runs while maintaining successful real-world locomotion performance after transfer.

In summary, this paper makes the following contributions:
\begin{enumerate}
    \item We present RANDPOL, a randomized policy learning approach for continuous control in which hidden layers are fixed after random initialization and only the final layer is optimized.
    \item We demonstrate that RANDPOL can learn an end-to-end locomotion controller for the Unitree Go2 quadruped and achieve successful zero-shot sim-to-real transfer.
    \item We show that, relative to PPO, RANDPOL attains comparative control performance with far fewer trainable parameters, a simpler optimization structure, and lower learning-phase computation time per iteration.
    \item We connect the empirical performance of RANDPOL to the mathematical foundation of randomized function approximation, providing a theory-backed perspective on why such reduced-parameter policy classes can still be effective.
\end{enumerate}

For reproducibility, code, hyperparameters, and hardware experiment videos are available at the anonymous project repository: \url{https://anonymous.4open.science/r/randpol-cdc-193F}.

\section{Related Work}
\label{sec:related_work}

\subsubsection{Random Features and Randomized Networks}
\label{subsec:rw_random_features}
A central idea behind this paper is that expressive nonlinear function classes can be constructed without training every parameter end-to-end. Random feature methods show that nonlinear models can be approximated by mapping inputs into a randomized feature space and learning only a linear readout \cite{c12, c13}. Related ideas also appear in randomized neural networks, reservoir computing, and deep randomized architectures, where hidden layers are fixed after random initialization and only a small subset of parameters is optimized \cite{c14, c18, c19, c20}. These works provide the main theoretical and conceptual motivation for RANDPOL.

\subsubsection{RL for Continuous Control}
\label{subsec:rw_continuous_control}
This paper is also related to reinforcement learning for continuous control. Modern policy-gradient and actor-critic methods, including PPO, TRPO, and related algorithms, typically rely on fully trainable neural network policies and value functions \cite{c2, c15, c3}. These methods have achieved strong empirical performance, but they usually optimize a large number of parameters and can be sensitive to architecture and hyperparameter choices. In contrast, our focus is on how much control performance can be retained when the trainable component is made much smaller.

\subsubsection{Quadruped Locomotion and Sim-to-Real}
\label{subsec:rw_locomotion}
Our work is further connected to learning-based locomotion control for legged robots. Prior work has demonstrated that deep reinforcement learning can produce effective quadruped locomotion policies in simulation and transfer them to hardware, often with strong robustness and adaptation capabilities \cite{c8, c9, c10, c11, c21, c22, c23, c24}. We follow this line of work in application domain, but study a different question: whether end-to-end locomotion control can remain effective when the policy class has drastically fewer trainable parameters.

\section{Preliminaries}
\label{sec:preliminaries}

We consider an infinite-horizon discounted Markov decision process (MDP)
\[
\mathcal{M} = (\mathcal{X}, \mathcal{U}, P, r, \gamma),
\]
where $\mathcal{X}$ is the state space, $\mathcal{U}$ is the action space, $P(\cdot \mid x,u)$ is the transition kernel, $r:\mathcal{X}\times\mathcal{U}\rightarrow\mathbb{R}$ is the reward function, and $\gamma \in (0,1)$ is the discount factor. Let $\Pi$ denote the class of stochastic Markov policies $\pi:\mathcal{X}\rightarrow\mathcal{P}(\mathcal{U})$.

For a trajectory $\tau=(x_0,u_0,x_1,u_1,\dots,x_T,u_T)$ generated by policy $\pi$, the discounted return is
\[
R(\tau)=\sum_{t=0}^{T}\gamma^t r(x_t,u_t),
\]
and the trajectory distribution is
\[
P(\tau\mid\pi)=\rho_0(x_0)\prod_{t=0}^{T-1}P(x_{t+1}\mid x_t,u_t)\pi(u_t\mid x_t),
\]
where $\rho_0$ is the initial-state distribution. The goal is to maximize the expected return
\[
J(\pi)=\mathbb{E}_{\tau\sim\pi}[R(\tau)],
\]
and obtain an optimal policy
\[
\pi^*=\arg\max_{\pi\in\Pi} J(\pi).
\]

Let $\pi_\theta$ be a stochastic policy parameterized by $\theta$. The policy-gradient theorem gives
\begin{equation}
\label{equation:policy_gradient}
\nabla_{\theta} J(\pi_\theta)
=
\mathbb{E}_{\tau\sim\pi_\theta}
\left[
\sum_{t=0}^{T}
\nabla_{\theta}\log \pi_\theta(u_t\mid x_t)\,G_t
\right],
\end{equation}
where
\[
G_t=\sum_{l=t}^{T}\gamma^{\,l-t}r(x_l,u_l)
\]
is the return-to-go from time step $t$. In actor-critic methods, $G_t$ is commonly replaced by an advantage estimate, often computed using generalized advantage estimation (GAE), to reduce gradient variance while preserving the overall policy-gradient structure \cite{c17}. 

RANDPOL is based on randomized neural networks, in which hidden-layer parameters are sampled once and then kept fixed while only a final linear readout is trained. This construction can be interpreted as learning a linear combination of random nonlinear basis functions. Prior results on random features and randomized neural networks show that such representations can approximate broad classes of functions, with approximation quality improving as the number of random features increases \cite{c12, c13, c14}. In this paper, these results motivate the use of fixed random features as compact yet expressive parameterizations for policies and value functions in continuous control.

\section{RANDPOL}
\label{sec:randpol}

We now introduce \emph{RANDomized POlicy Learning} (RANDPOL), a policy learning approach in which both the actor and the critic are parameterized by randomized neural networks. Instead of optimizing all parameters of a deep network, RANDPOL samples the hidden layers once, keeps them fixed throughout training, and optimizes only the final linear layer. This drastically reduces the number of trainable parameters while preserving nonlinear expressiveness through the fixed random basis.

\subsection{Randomized Actor and Critic}
\label{subsec:randpol_param}

Let $\phi(\cdot;\theta_j):\mathcal{X}\rightarrow\mathbb{R}$, for $j=1,\dots,J_Q$, denote random basis functions generated by a frozen randomized neural network for the critic. The value function is approximated by
\begin{equation}
\label{equation:critic_randpol}
V_{\alpha}(x)=\sum_{j=1}^{J_Q}\alpha_j \phi(x;\theta_j),
\end{equation}
where $\alpha\in\mathbb{R}^{J_Q}$ is the trainable readout weight vector.

Similarly, let $\psi(\cdot;\vartheta_j):\mathcal{X}\rightarrow\mathbb{R}^{m}$, for $j=1,\dots,J_\pi$, denote random basis functions generated by a frozen randomized policy network, where $m=\dim(\mathcal{U})$. The policy mean is parameterized as
\begin{equation}
\label{equation:actor_mean_randpol}
\mu_{\beta}(x)=\sum_{j=1}^{J_\pi}\beta_j \psi(x;\vartheta_j),
\end{equation}
where $\beta$ is the trainable readout parameter. For continuous control, we use a Gaussian policy
\begin{equation}
\label{equation:actor_policy_randpol}
\pi_{\beta}(u\mid x)=\mathcal{N}\!\left(\mu_{\beta}(x),\Sigma\right),
\end{equation}
where $\Sigma$ is the policy covariance used in implementation.

Under this construction, both the actor and the critic are linear models on top of fixed random nonlinear features. RANDPOL therefore retains a nonlinear policy and value class while greatly shrinking the trainable parameter space.

\subsection{Practical Training Objective}
\label{subsec:randpol_training}

In implementation, RANDPOL follows the same rollout-based training paradigm as modern on-policy actor-critic methods, but restricts learning to the final readout of the actor and critic. Given a rollout
\[
\{(x_t,u_t,r_t,x_{t+1})\}_{t=0}^{T},
\]
we first estimate advantages and value targets using generalized advantage estimation. The critic readout is then fit on the fixed random features by minimizing the squared error
\begin{equation}
\label{equation:critic_loss}
\mathcal{L}_V(\alpha)
=
\frac{1}{N_Q}\sum_{n=1}^{N_Q}
\left(V_{\alpha}(x_n)-\hat{V}_n\right)^2,
\end{equation}
where $\hat{V}_n$ denotes the target value estimate and $\{x_n\}_{n=1}^{N_Q}$ are sampled states from the rollout.

For the actor, we optimize only the readout parameter $\beta$ while keeping the random basis fixed. Let $\hat{A}_t$ denote the estimated advantage at time step $t$. Using the probability ratio
\[
r_t(\beta)=\frac{\pi_{\beta}(u_t\mid x_t)}{\pi_{\beta_{\mathrm{old}}}(u_t\mid x_t)},
\]
we use the clipped surrogate objective
\begin{equation}
\label{equation:actor_loss}
\mathcal{L}_{\pi}(\beta)
=
-\frac{1}{N_\pi}\sum_{n=1}^{N_\pi}
\min\!\left(
r_n(\beta)\hat{A}_n,\;
\mathrm{clip}(r_n(\beta),1-\epsilon,1+\epsilon)\hat{A}_n
\right),
\end{equation}
where $\epsilon>0$ is the clipping parameter and $\{(x_n,u_n)\}_{n=1}^{N_\pi}$ are sampled state-action pairs from the rollout. In this way, RANDPOL uses a stable on-policy training objective while preserving the low-dimensional optimization problem induced by the randomized parameterization.

This practical training setup keeps the optimization pipeline close to standard on-policy actor-critic methods, while RANDPOL differs primarily in its randomized low-dimensional actor-critic parameterization. 

The complete training procedure is summarized in Algorithm~\ref{alg:randpol}.

\begin{algorithm}[t]
\caption{\textbf{RANDPOL:} Randomized Policy Learning}
\label{alg:randpol}
\begin{algorithmic}[1]
\State \textbf{Input:} fixed random feature parameters $\{\theta_j\}_{j=1}^{J_Q}$ and $\{\vartheta_j\}_{j=1}^{J_\pi}$, initial readout weights $\alpha,\beta$
\Repeat
    \State Collect rollout data using policy $\pi_\beta$
    \State Compute value targets and advantage estimates from the rollout
    \State Update critic readout $\alpha$ by minimizing \eqref{equation:critic_loss}
    \State Update actor readout $\beta$ by minimizing \eqref{equation:actor_loss} for multiple epochs over minibatches
\Until{convergence}
\end{algorithmic}
\end{algorithm}

\subsection{Design Perspective}
\label{subsec:randpol_discussion}

RANDPOL serves as a parameter-efficient alternative for studying a different trade-off in learning-based control. If the task is structured enough, a rich fixed random basis together with a trainable linear readout may already be sufficient to achieve effective control behavior. This design reduces trainable complexity, simplifies optimization, and can lead to more stable learning behavior, while still preserving effective sim-to-real performance in practice.

From this perspective, the contribution of RANDPOL is not only a randomized actor-critic parameterization, but also a concrete demonstration that competitive end-to-end locomotion control can be achieved within a much smaller trainable parameter space. This performance-complexity trade-off is central to the empirical study in the following section.

\section{Experiments}
\label{sec:experiments}

We evaluate RANDPOL on end-to-end quadruped locomotion with the Unitree Go2 robot. Our experiments are designed to answer three questions. First, how close can RANDPOL get to PPO on a locomotion tracking task in simulation? Second, what is the trade-off between control performance and trainable complexity? Third, can a RANDPOL controller trained in simulation transfer zero-shot to the physical robot?

\subsection{Experimental Setup}
\label{subsec:exp_setup}

\subsubsection{Simulation Environment and Task}
\label{subsubsec:sim_env}

All experiments are conducted in NVIDIA Isaac Lab using the Unitree Go2 quadruped \cite{c16}. The simulator uses a physics time step of $0.005$~s and a control decimation of $4$, corresponding to a control frequency of $50$~Hz. Each training episode lasts $20$~s, and all simulation experiments use $4096$ parallel environments.

\begin{figure}[t]
    \centering
    \begin{subfigure}[t]{0.48\linewidth}
        \centering
        \includegraphics[width=\linewidth]{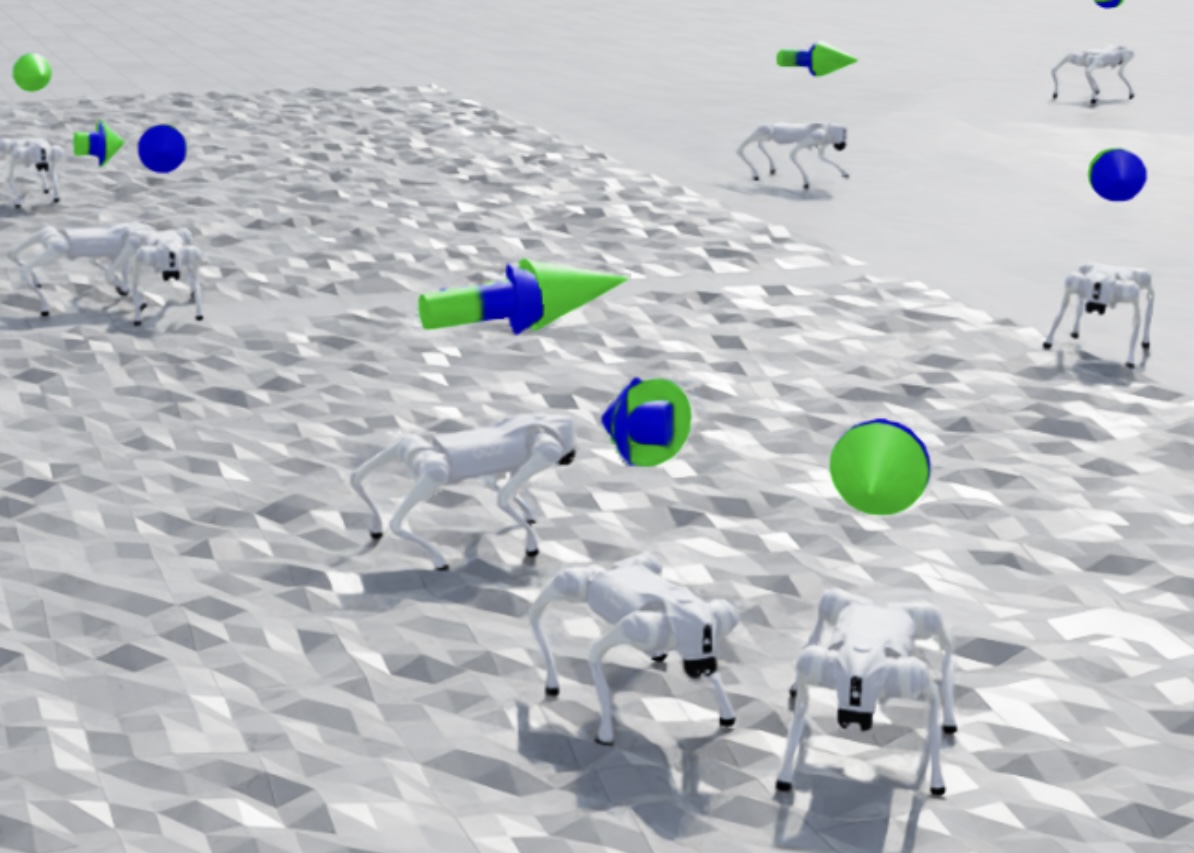}
        \caption{Rough terrain in simulation.}
    \end{subfigure}\hfill
    \begin{subfigure}[t]{0.48\linewidth}
        \centering
        \includegraphics[width=\linewidth]{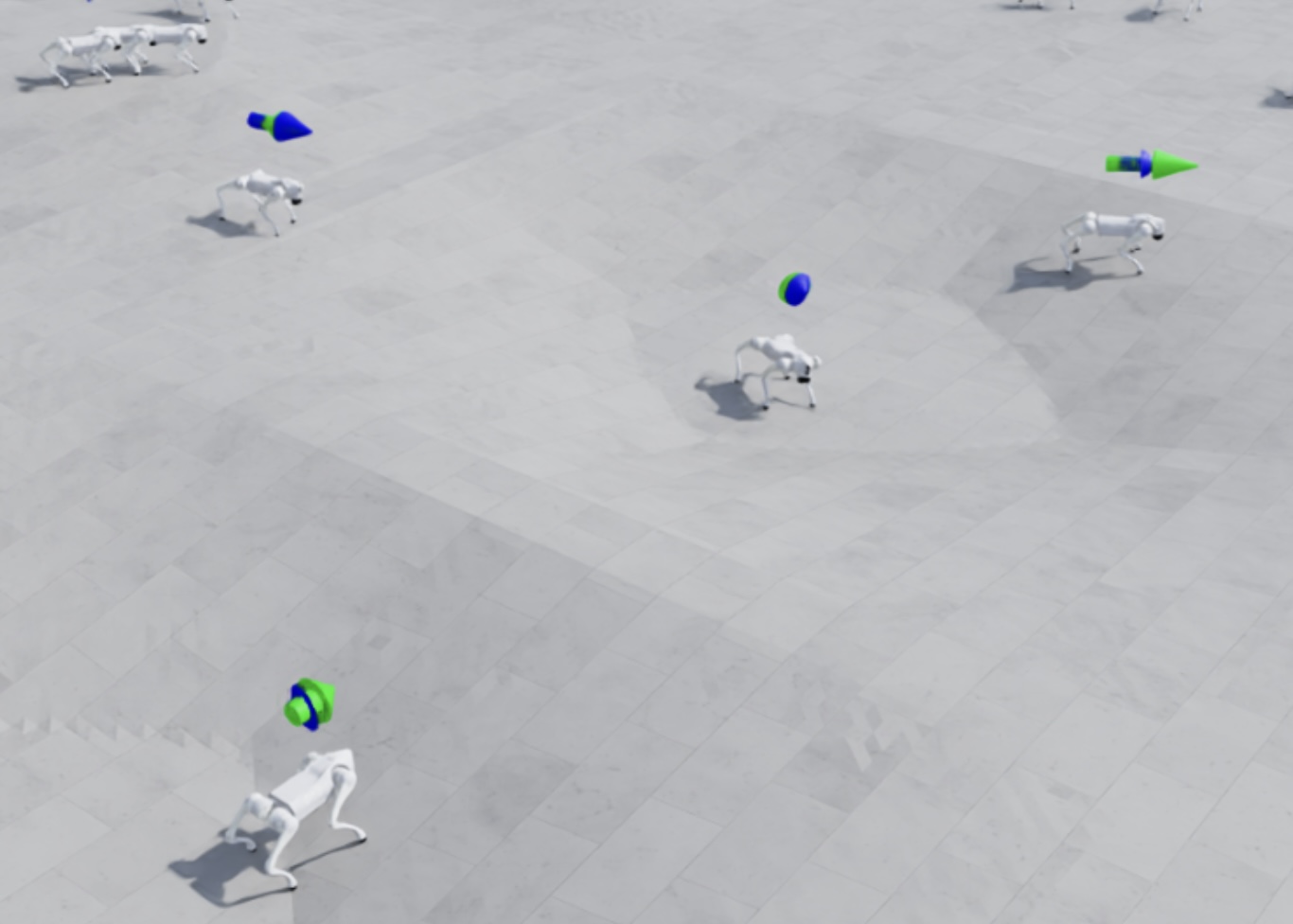}
        \caption{Uphill and downhill terrain regions in simulation.}
    \end{subfigure}
    \caption{Representative simulation environments used for training. The terrain setup covers flat ground, rough terrain, and upward and downward slopes, supporting curriculum-based locomotion training under diverse surface conditions.}
    \label{fig:sim_terrain}
\end{figure}

The control task is body-velocity tracking for the Unitree Go2. During training, the robot is commanded to track a desired forward linear velocity and yaw angular velocity, with commands resampled every $10$~s. The command ranges are initialized as
\begin{equation}
v_x \in [0.0,\,0.2]~\mathrm{m/s}, \qquad
\omega_z \in [-0.2,\,0.2]~\mathrm{rad/s},
\end{equation}
and are gradually expanded to
\begin{equation}
v_x \in [0.0,\,1.0]~\mathrm{m/s}, \qquad
\omega_z \in [-1.0,\,1.0]~\mathrm{rad/s}.
\end{equation}
The command curriculum is promotion-based: the command limits are expanded only when the episode-averaged tracking performance exceeds a predefined threshold. In addition to the command curriculum, training is performed on procedurally generated terrain with a terrain curriculum spanning flat ground, rough terrain, and upward and downward slopes, as illustrated in Fig.~\ref{fig:sim_terrain}.

The policy outputs $12$ joint position commands, one for each actuated joint of Go2. These commands are interpreted as offsets around the nominal standing configuration and are scaled by $0.25$ before being tracked by the robot's low-level joint controllers. The policy observation is purely proprioceptive and includes base angular velocity, projected gravity, the commanded velocity, joint position errors, joint velocities, and the previous action. The critic additionally receives privileged information, including base linear velocity and joint effort. No observation history is used in the present Go2 experiments.

To improve robustness, the simulation includes domain randomization during training. Specifically, rigid-body friction and restitution are randomized, the base mass is perturbed, the robot base pose is randomized at reset, joint velocities are randomized at reset, and random pushes are applied during training. These perturbations are enabled in simulation only and are not used in the hardware experiment.

\subsubsection{Learning Setup}
\label{subsubsec:learning_setup}

We compare RANDPOL with PPO under the same simulation environment, reward function, and curriculum design. This comparison is intended to isolate the effect of policy parameterization and trainable complexity rather than differences in task specification.

For RANDPOL, both the actor and critic use frozen random nonlinear features with trainable linear readouts. The current implementation uses a random feature dimension of $400$, one hidden random layer of width $500$, ELU activation, and uniform random initialization. Training uses $50$ environment steps per iteration, discount factor $\gamma=0.99$, and GAE parameter $\lambda=0.95$. The policy is optimized with a clipped surrogate objective for $5$ epochs and $4$ minibatches per iteration. The policy learning rate is $3\times 10^{-4}$, the entropy coefficient is $0.01$, and gradient clipping is set to $0.5$. Observation normalization, reward normalization, and advantage normalization are enabled.

For PPO, we use separate multilayer perceptrons for the actor and critic, each with hidden dimensions $(512,256,128)$ and ELU activation. PPO uses $24$ environment steps per iteration, $5$ learning epochs, and $4$ minibatches. The learning rate is $10^{-3}$ with adaptive scheduling based on the KL divergence. The clipping parameter is $0.2$, the entropy coefficient is $0.01$, the discount factor is $\gamma=0.99$, and the GAE parameter is $\lambda=0.95$. The PPO baseline is implemented with \texttt{rsl\_rl}, following the official Unitree RL Lab repository \cite{c25}.

For both methods, the training horizon is $1000$ iterations. All reported simulation results are aggregated over five runs for each algorithm.

\subsubsection{Evaluation Protocol}
\label{subsubsec:eval_protocol}

We evaluate the methods along three axes: simulation performance, learning-phase efficiency, and hardware tracking performance.

In simulation, we report mean episodic reward, linear-velocity tracking reward, yaw-rate tracking reward, forward-velocity tracking error, and yaw-rate tracking error. These metrics distinguish overall return from actual command-tracking quality.

To study computational efficiency, we report aggregated learning-phase timing rather than total wall-clock time, since RANDPOL and PPO use different rollout lengths. We also report trainable parameter counts and total parameter counts for both methods.

On hardware, we evaluate RANDPOL on the physical Unitree Go2. The hardware study focuses on forward-velocity and yaw-rate tracking under user-issued commands. We compare commanded and measured robot velocities using both time-series plots and scalar tracking errors.

\subsection{Results}
\label{subsec:results}

\subsubsection{Simulation Performance}
\label{subsubsec:sim_results}

\begin{figure*}[t]
    \centering
    \begin{subfigure}[t]{0.32\textwidth}
        \centering
        \includegraphics[width=\linewidth]{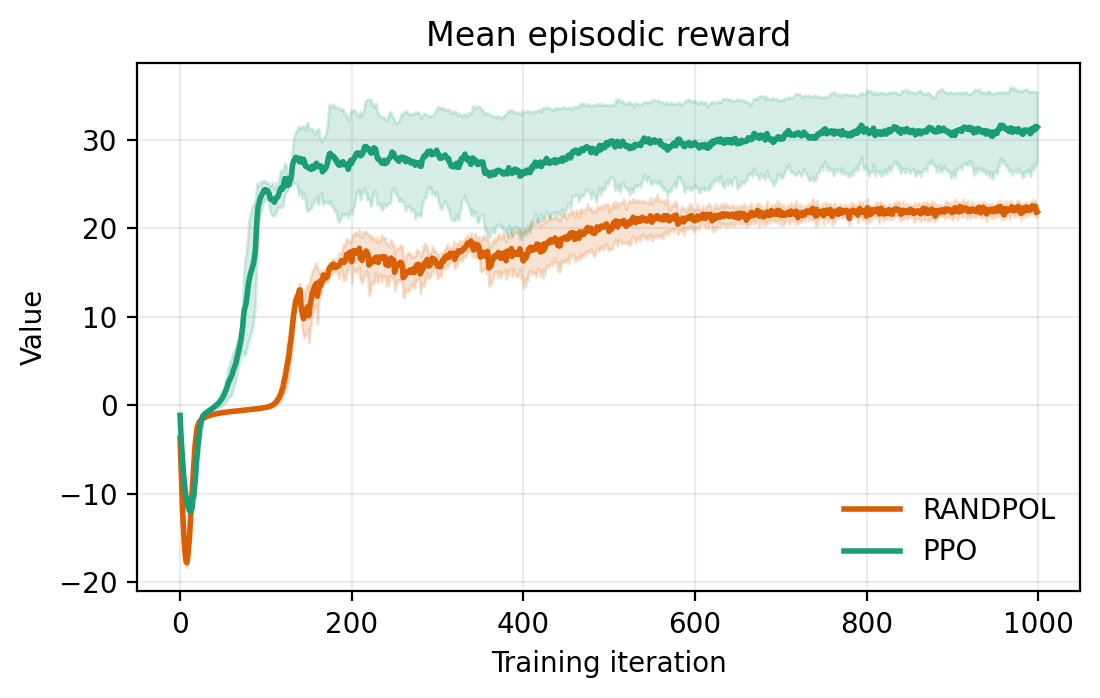}
        \caption{Mean episodic reward.}
    \end{subfigure}\hfill
    \begin{subfigure}[t]{0.32\textwidth}
        \centering
        \includegraphics[width=\linewidth]{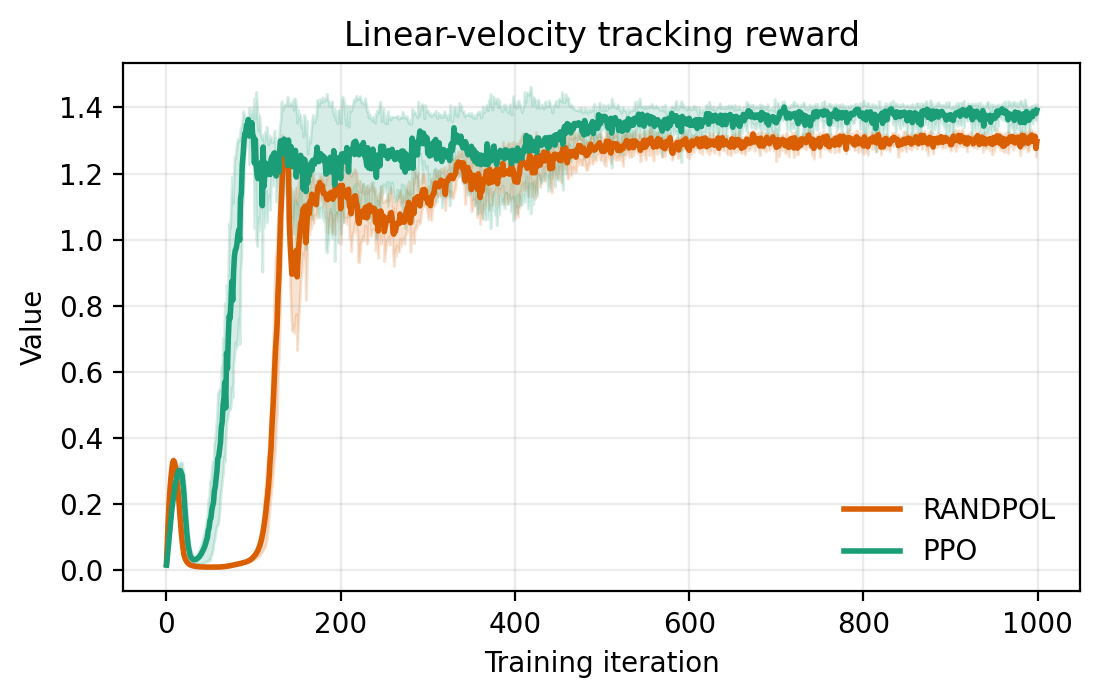}
        \caption{Linear-velocity tracking reward.}
    \end{subfigure}\hfill
    \begin{subfigure}[t]{0.32\textwidth}
        \centering
        \includegraphics[width=\linewidth]{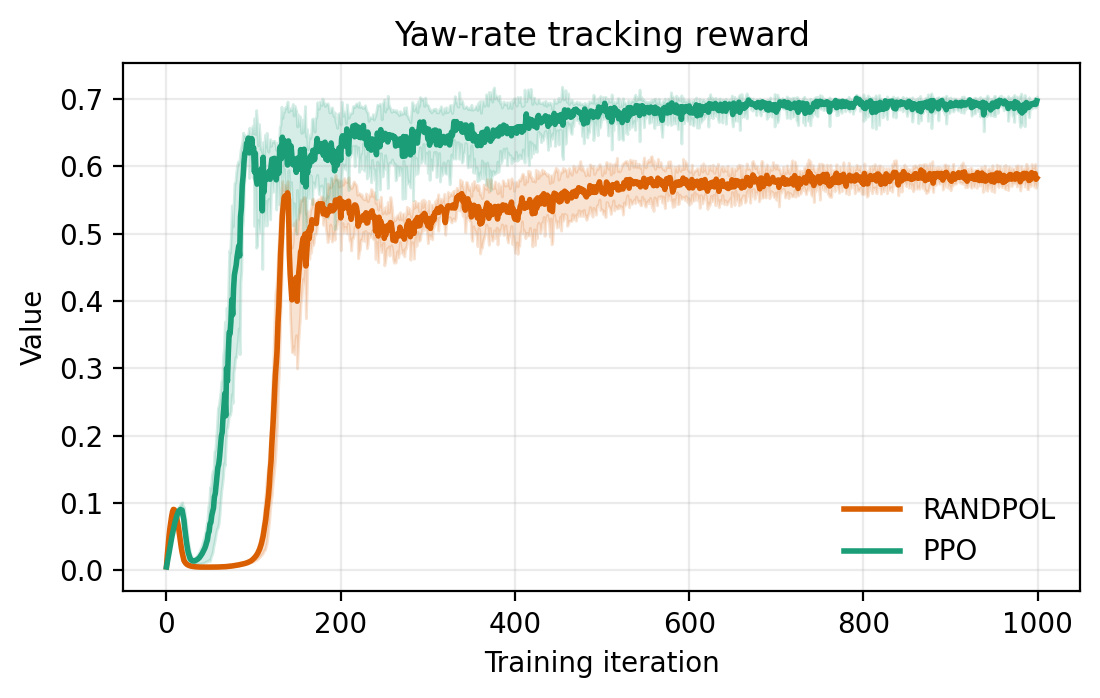}
        \caption{Yaw-rate tracking reward.}
    \end{subfigure}
    \caption{Training curves on the Go2 forward-and-yaw velocity-tracking task. Each plot shows the mean across five runs together with a 95\% confidence interval.}
    \label{fig:sim_learning_curves}
\end{figure*}

Figure~\ref{fig:sim_learning_curves} summarizes the training behavior of RANDPOL and PPO on the Go2 forward-and-yaw velocity-tracking task. Each curve shows the mean across five runs together with a 95\% confidence interval. Table~\ref{tab:sim_results} reports the final simulation performance, also aggregated over five runs.

Overall, PPO achieves higher final reward and lower tracking errors than RANDPOL on this task. However, RANDPOL still attains comparative locomotion performance despite using a much smaller trainable parameter set. In particular, RANDPOL reaches a final mean reward of $21.834 \pm 0.595$, compared with $31.416 \pm 3.936$ for PPO. RANDPOL remains relatively close to PPO in the forward-velocity tracking term and forward-velocity error, while the performance gap is larger in yaw tracking and the associated yaw-rate error.

These results should be interpreted together with the complexity comparison in Table~\ref{tab:learning_time}. The key point of this paper is not that RANDPOL surpasses PPO in absolute performance, but that it preserves useful locomotion capability while drastically reducing the dimension of the optimization problem. The five-run results also show that RANDPOL exhibits noticeably lower variability in mean reward than PPO in the current setup.

\begin{table}[t]
    \centering
    \caption{Simulation results on the Go2 forward-and-yaw velocity-tracking task, aggregated over five runs per algorithm.}
    \label{tab:sim_results}
    \setlength{\tabcolsep}{12pt}
    \begin{tabular}{lcc}
        \toprule
        Metric & RANDPOL & PPO \\
        \midrule
        Mean reward & $21.834 \pm 0.595$ & $31.416 \pm 3.936$ \\
        Linear tracking reward & $1.297 \pm 0.016$ & $1.392 \pm 0.018$ \\
        Yaw tracking reward & $0.583 \pm 0.009$ & $0.697 \pm 0.007$ \\
        Forward-velocity error & $0.266 \pm 0.005$ & $0.225 \pm 0.009$ \\
        Yaw-rate error & $0.394 \pm 0.024$ & $0.215 \pm 0.004$ \\
        \bottomrule
    \end{tabular}
\end{table}

\subsubsection{Learning-Time and Complexity Comparison}
\label{subsubsec:time_results}

Table~\ref{tab:learning_time} reports the aggregated learning time for RANDPOL and PPO. The purpose of this comparison is not to measure simulator throughput, but to compare the computational cost of the two learning procedures. RANDPOL uses frozen nonlinear features with trainable linear readouts, whereas PPO performs end-to-end gradient-based updates of both actor and critic networks.

A central observation is that RANDPOL reduces the trainable parameter count dramatically, from $377{,}241$ in PPO to $5{,}225$. This is the main efficiency axis of interest in this paper. At the same time, RANDPOL has a larger total parameter count because the fixed random basis is retained as part of the model. This distinction between total parameters and trainable parameters is important: RANDPOL does not seek to minimize model size in an absolute sense, but rather to reduce the dimension of the optimization problem.

In terms of learning-phase timing, RANDPOL and PPO are close in the current implementation, with RANDPOL showing a slightly lower total learning time per iteration. Taken together with the simulation results, these numbers support the central trade-off studied in this paper: RANDPOL gives up some performance relative to PPO, but does so while optimizing a much smaller trainable model.

\begin{table}[t]
    \centering
    \caption{Learning-phase timing and complexity comparison.}
    \label{tab:learning_time}
    \setlength{\tabcolsep}{5.8pt}
    \begin{tabular}{lcc}
        \toprule
        Metric & RANDPOL & PPO \\
        \midrule
        Total learning time per iteration & $0.0563 \pm 0.0003$ & $0.0747 \pm 0.0005$ \\
        Trainable parameter count & 5,225 & 377,241 \\
        Total parameter count & 459,525 & 377,241 \\
        \bottomrule
    \end{tabular}
\end{table}

\subsubsection{Hardware Tracking Results}
\label{subsubsec:hardware_results}

We deploy the learned RANDPOL controller on the physical Unitree Go2 and evaluate its real-world behavior under user-issued commands. To execute the learned policy, the robot's built-in sports mode must be disabled. In this mode, the base-velocity estimate provided by the default control stack is no longer available. As a result, our quantitative hardware evaluation focuses on forward-velocity tracking, for which external measurements are available, while yaw-rate tracking is assessed qualitatively from the robot response trajectories.

\begin{figure}[t]
    \centering
    \begin{subfigure}[t]{0.24\linewidth}
        \centering
        \includegraphics[width=\linewidth]{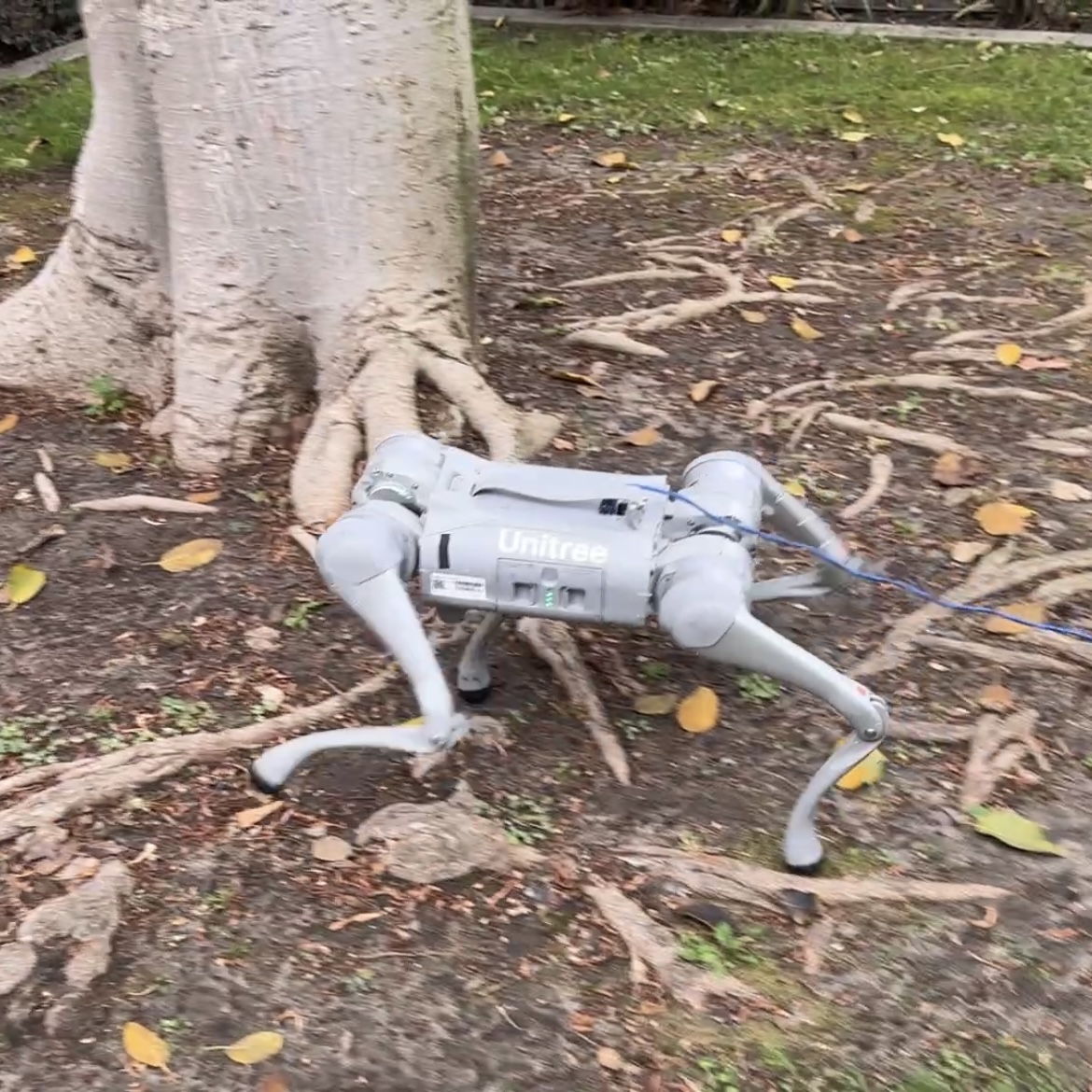}
        \caption{Grass}
    \end{subfigure}\hfill
    \begin{subfigure}[t]{0.24\linewidth}
        \centering
        \includegraphics[width=\linewidth]{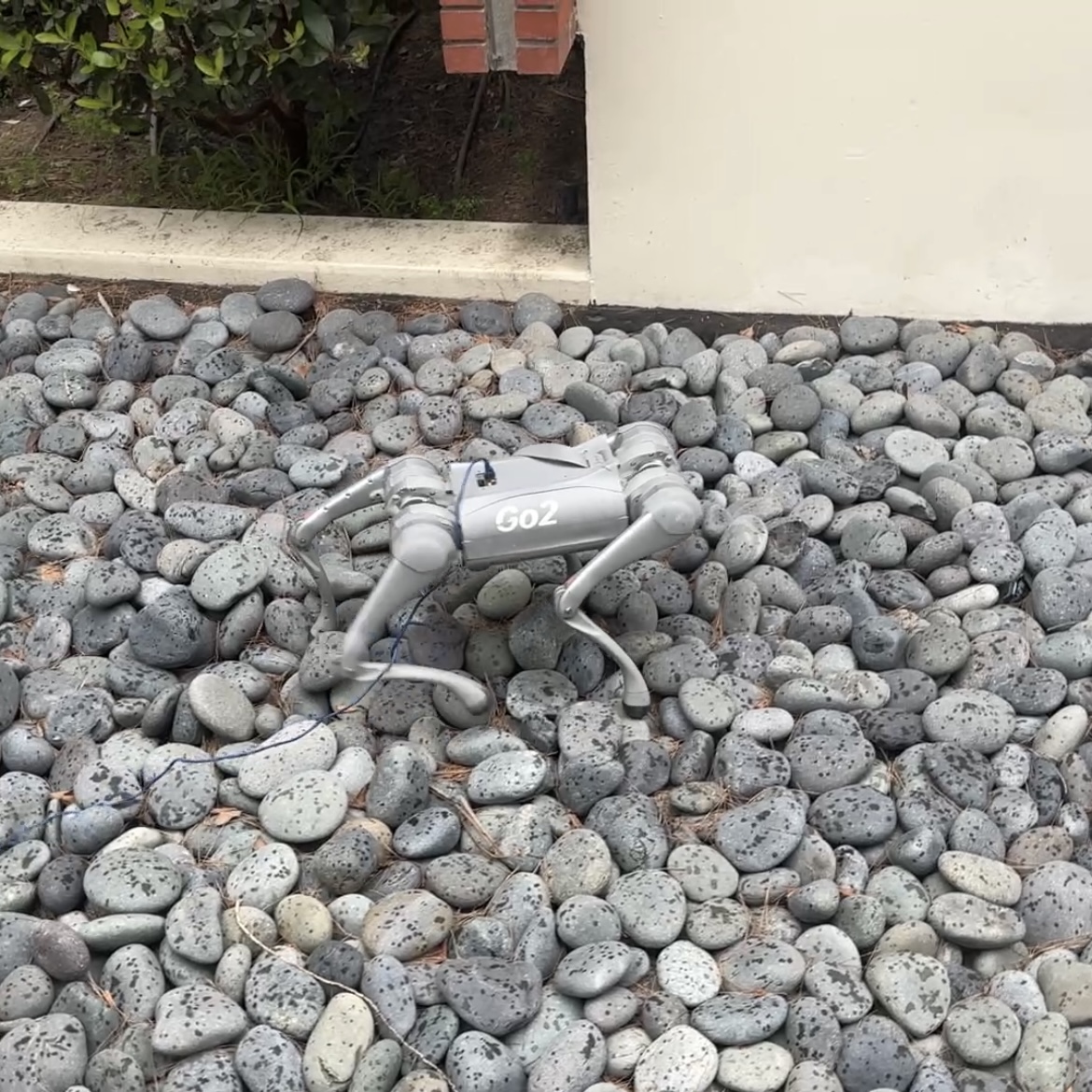}
        \caption{Stone}
    \end{subfigure}\hfill
    \begin{subfigure}[t]{0.24\linewidth}
        \centering
        \includegraphics[width=\linewidth]{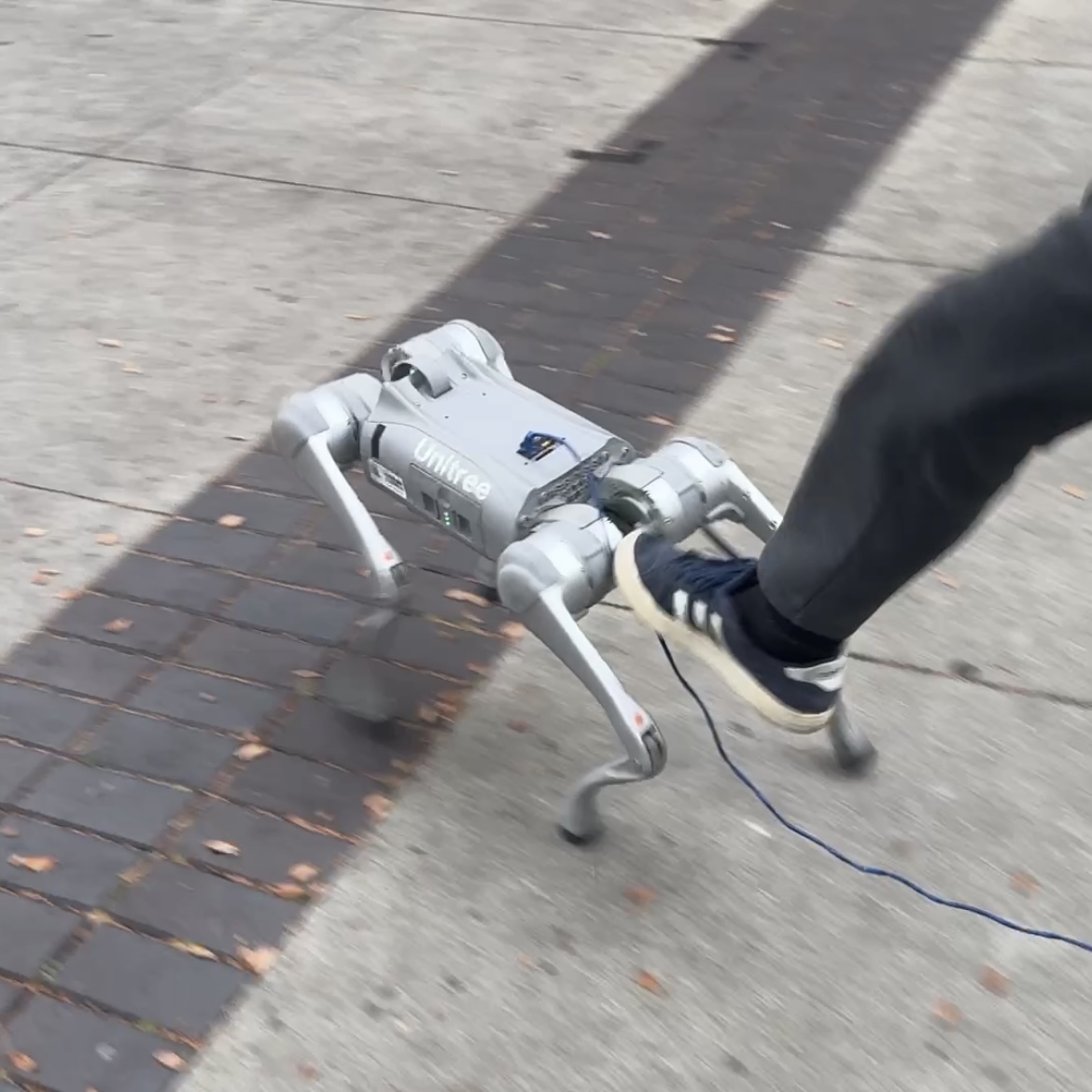}
        \caption{Kick}
    \end{subfigure}\hfill
    \begin{subfigure}[t]{0.24\linewidth}
        \centering
        \includegraphics[width=\linewidth]{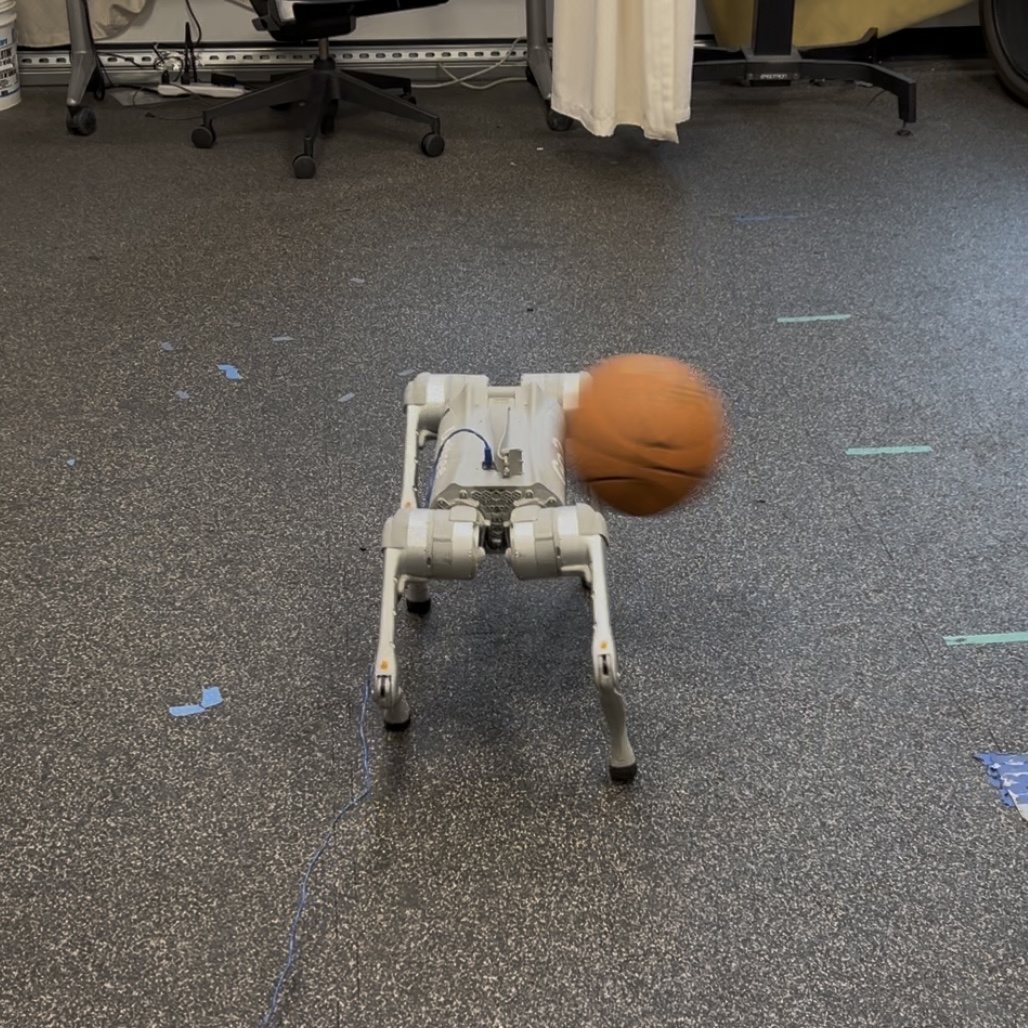}
        \caption{Ball hit}
    \end{subfigure}
    \caption{Representative real-world evaluation conditions for RANDPOL on the Unitree Go2: bumpy grass, uneven stone surfaces, and external disturbances including kicking and basketball impact. These examples illustrate that the hardware evaluation is intentionally challenging.}
    \label{fig:hardware_conditions}
\end{figure}

\begin{figure}[t]
    \centering
    \includegraphics[width=\linewidth]{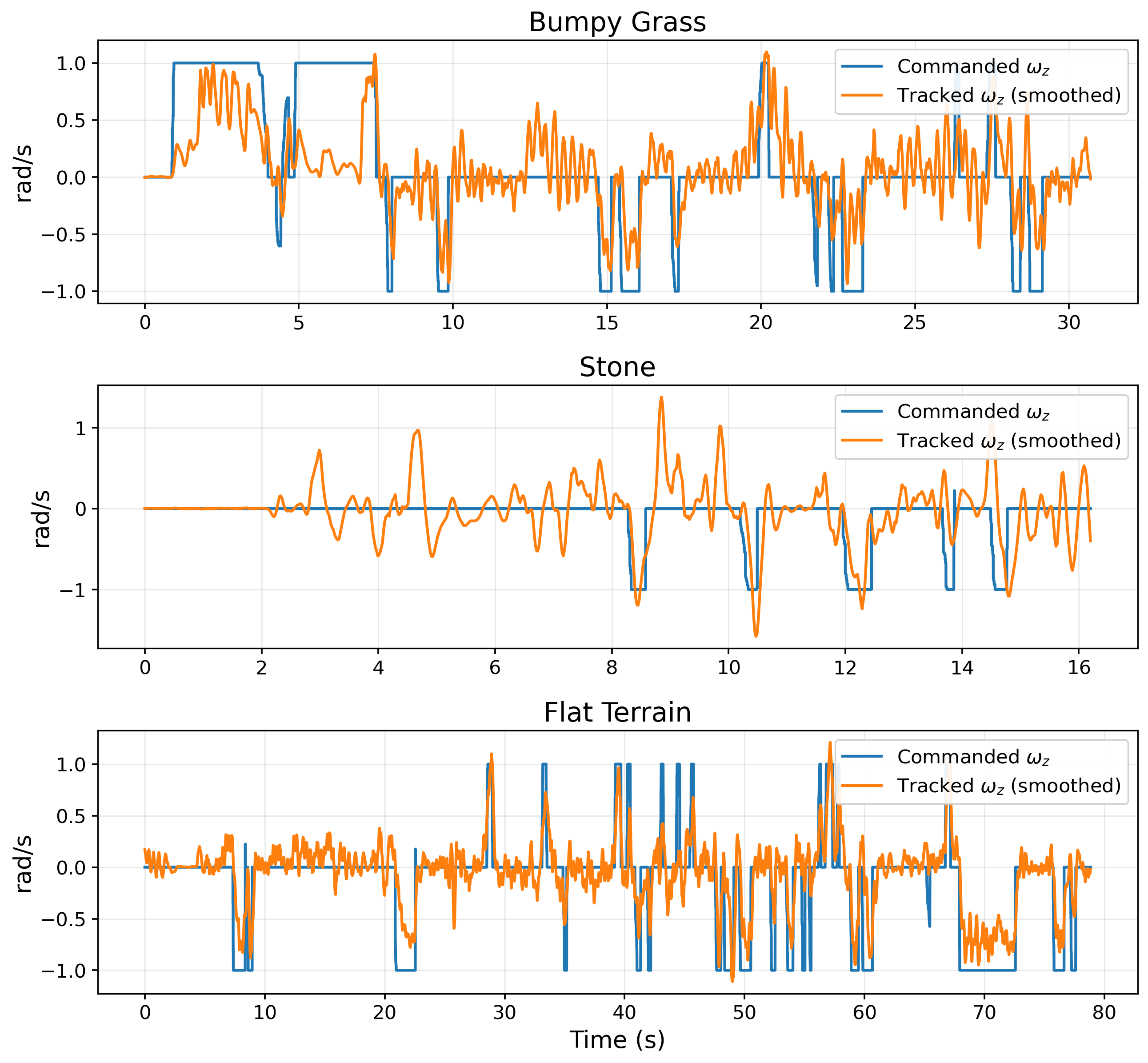}
    \caption{Representative hardware yaw-rate tracking trajectories for RANDPOL on the Unitree Go2 across multiple terrains. From top to bottom: bumpy grass, stone, and flat terrain. The plots show user-issued yaw-rate commands and the corresponding tracked response after deployment on hardware.}
    \label{fig:hardware_tracking}
\end{figure}

The hardware experiments are conducted on multiple terrains, including flat terrain, bumpy grass, and stone, and also include external disturbance tests, as shown in Fig.~\ref{fig:hardware_conditions}. Figure~\ref{fig:hardware_tracking} shows representative yaw-rate tracking trajectories across multiple terrains, illustrating that the learned controller remains responsive to user-issued turning commands beyond the nominal flat-ground setting. For quantitative evaluation, however, we report forward-velocity tracking errors on flat terrain only. This distinction is important because terrain-induced speed deviations vary substantially across surfaces, so aggregating MAE or RMSE across heterogeneous terrains would confound controller tracking quality with terrain difficulty.

The hardware results demonstrate successful zero-shot sim-to-real transfer of the learned RANDPOL controller. The robot exhibits stable locomotion and responsive turning behavior across the tested terrains without additional real-world fine-tuning, showing that the parameter-efficient policy learned in simulation remains effective after deployment on the physical platform.

\begin{table}[t]
    \centering
    \caption{Hardware forward-velocity tracking errors for RANDPOL on Unitree Go2, reported on flat terrain.}
    \label{tab:hardware_results}
    \begin{tabular}{lc}
        \toprule
        Metric & RANDPOL \\
        \midrule
        Forward-velocity MAE & 0.2602 \\
        Forward-velocity RMSE & 0.3697 \\
        \bottomrule
    \end{tabular}
\end{table}

\section{Conclusions and Future Work}
\label{sec:conclusion}

\subsection{Conclusions}
\label{subsec:conclusions}

This paper presented RANDPOL for end-to-end quadruped locomotion control with a highly reduced trainable parameter space. The central idea is to use a rich fixed random nonlinear basis together with a trainable linear readout, yielding a controller that is substantially simpler to optimize while still maintaining effective control behavior.

On the Unitree Go2 locomotion task, RANDPOL achieved comparative body-velocity tracking performance in simulation while using dramatically fewer trainable parameters than PPO. In our implementation, RANDPOL reduces the number of trainable parameters from $377{,}241$ to $5{,}225$, while attaining a final mean reward of $21.834 \pm 0.595$ compared with $31.416 \pm 3.936$ for PPO. The aggregated learning time per iteration is also slightly lower for RANDPOL in the current setup. These results highlight a clear performance-complexity trade-off: although PPO remains stronger in absolute reward and tracking accuracy, RANDPOL preserves useful locomotion capability within a much lower-dimensional optimization problem.

The hardware experiments further confirm the practical value of this design. The learned RANDPOL controller transfers zero-shot from simulation to the physical Unitree Go2 and achieves successful forward-velocity and yaw-rate tracking under user-issued commands. This demonstrates that the proposed parameter-efficient policy class is not only effective in simulation, but also capable of producing stable and usable real-world locomotion behavior after transfer.

Overall, the results establish RANDPOL as a practical and principled approach for parameter-efficient locomotion control. The broader takeaway is that, in structured robotic control problems, reducing trainable complexity can remain compatible with effective simulated and real-world performance.

\subsection{Future Work}
\label{subsec:future_work}

Several directions remain open for future study. First, it would be valuable to evaluate RANDPOL on a broader range of locomotion tasks, including richer command spaces, more aggressive maneuvers, and more challenging terrain conditions, to better understand the scope of the observed performance-complexity trade-off.

Second, future work could explore alternative random feature constructions, hidden-layer widths, and initialization strategies in order to better characterize how the richness of the fixed basis affects control performance. Since RANDPOL relies on fixed random nonlinear features, understanding these design choices more systematically may lead to further gains in efficiency and tracking quality.

Third, a broader empirical study of robustness, variance, and hyperparameter sensitivity would help clarify how consistently the practical simplicity of RANDPOL carries across tasks and training conditions.

Finally, deeper theoretical analysis connecting randomized function approximation to closed-loop control performance may provide further insight into why reduced-parameter policy classes can work effectively in structured robotic control problems.





\end{document}